\documentclass[conference]{IEEEtran}

\ifCLASSINFOpdf

\usepackage{graphicx}
\usepackage{xcolor}
\usepackage[scriptsize,tight]{subfigure}
\usepackage{tabularx}
	\newcolumntype{C}[1]{>{\centering\arraybackslash}p{#1}}
	\newcolumntype{R}[1]{>{\raggedleft\arraybackslash}p{#1}}
	\newcolumntype{L}[1]{>{\raggedright\arraybackslash}p{#1}}
\usepackage{booktabs}	
\usepackage[ruled,linesnumbered]{algorithm2e}
\usepackage{amssymb}
\usepackage{xspace}
\newcommand{\R}{\mbox{$\mathbb{R}$}}

\begin{document}
%\title{Non-Linear Dimensionality Reduction by Unsupervised K-Nearest-Neighbor Regression}
\title{Unsupervised K-Nearest Neighbor Regression}


\author{\IEEEauthorblockN{Oliver Kramer}
%\author{\IEEEauthorblockN{}
\IEEEauthorblockA{Fakult\"{a}t II, Department f\"{u}r Informatik\\
Carl von Ossietzky Universit\"{a}t Oldenburg\\
26111 Oldenburg, Germany\\
\textit{oliver.kramer@uni-oldenburg.de}}
%}
}


%Telephone: (800) 555--1212, Fax: (888) 555--1212}


\maketitle

\begin{abstract}
In many scientific disciplines structures in high-dimensional data have to be found, e.g., in stellar spectra, in genome data, or in face recognition tasks. In this work we present a novel approach to non-linear dimensionality reduction. It is based on fitting K-nearest neighbor regression to the unsupervised regression framework for learning of low-dimensional manifolds. Similar to related approaches that are mostly based on kernel methods, unsupervised K-nearest neighbor (UNN) regression optimizes latent variables w.r.t. the data space reconstruction error employing the K-nearest neighbor heuristic. The problem of optimizing latent neighborhoods is difficult to solve, but the UNN formulation allows the design of efficient strategies that iteratively embed latent points to fixed neighborhood topologies. UNN is well appropriate for sorting of high-dimensional data. The iterative variants are analyzed experimentally.
\end{abstract}


% \fi
\IEEEpeerreviewmaketitle

\section{Introduction}

Dimensionality reduction and manifold learning have an important part to play in the understanding of data. In this work we introduce two fast constructive heuristics for dimensionality reduction called unsupervised K-nearest neighbor regression. Meinicke \cite{Meinicke} proposed a general unsupervised regression framework for learning of low-dimensional manifolds. The idea is to reverse the regression formulation such that low-dimensional data samples in latent space optimally reconstruct high-dimensional output data. We take this framework as basis for an iterative approach that fits KNN to this unsupervised setting in a combinatorial variant.
The manifold problem we consider is a mapping $\mathbf{F}: \mathbf{y} \rightarrow \mathbf{x}$ corresponding to the dimensionality reduction for data points $\mathbf{y} \in \mathbf{Y} \subset \R^d$, and latent points $\mathbf{x} \in \mathbf{Y} \subset \R^q$ with $d > q$. The problem is a hard optimization problem as the latent variables $\mathbf{X}$ are unknown. 

In Section \ref{sec:related} we will review related dimensionality reduction approaches, and repeat KNN regression. Section \ref{sec:UNN} presents the concept of UNN regression, and two iterative strategies that are based on fixed latent space topologies. Conclusions are drawn in Section \ref{sec:conclusions}.

\section{Related Work}
\label{sec:related}

Many dimensionality reduction methods have been proposed, a very famous one is principal component analysis (PCA), which assumes linearity of the manifold \cite{pca,pearson}. An extension for learning of non-linear manifolds is kernel PCA \cite{kernelpca} that projects the data into a Hilbert space. Further famous approaches for manifold learning are Isomap by Tenenbaum, Silva, and Langford \cite{isomap}, locally linear embedding (LLE) by Roweis and Saul \cite{lle}, and principal curves by Hastie and Stuetzle \cite{principal}.

\subsection{Unsupervised Regression}

The work on unsupervised regression for dimensionality reduction starts with Meinicke \cite{Meinicke}, who introduced the corresponding algorithmic framework for the first time. In this line of research early work concentrated on non-parametric kernel density regression, i.e., the counterpart of the Nadaraya-Watson estimator \cite{ritter05} denoted as unsupervised kernel regression (UKR). Klanke and Ritter \cite{ritter07} introduced an optimization scheme based on LLE, PCA, and leave-one-out cross-validation (LOO-CV) for UKR. Carreira-Perpi{\~n}{\'a}n and Lu \cite{gilroy} argue that training of non-parametric unsupervised regression approaches is quite expensive, i.e., $\mathcal{O}(N^3)$ in time, and $\mathcal{O}(N^2)$ in memory. Parametric methods can accelerate learning, e.g., unsupervised regression based on radial basis function networks (RBFs) \cite{regularized}, Gaussian processes \cite{lawrence}, and neural networks \cite{tan}.

\subsection{KNN Regression}
\label{KNN Regression}

In the following, we give a short introduction to K-nearest neighbor regression that is basis of the UNN approach. The problem in regression is to predict output values $\mathbf{y} \in \R^d$ to given input values $\mathbf{x} \in \R^q$ based on sets of $N$ input-output examples $((\mathbf{x}_1,\mathbf{y}_1),\ldots,(\mathbf{x}_N,\mathbf{y}_N))$. The goal is to learn a function $\mathbf{f}: \mathbf{x} \rightarrow \mathbf{y}$ known as regression function. We assume that a data set consisting of observed pairs $(\textbf{x}_i, \textbf{y}_i) \in \mathbf{X} \times \mathbf{Y}$ is given.
For a novel pattern $\mathbf{x}'$, KNN regression computes the mean of the function values of its K-nearest neighbors:
\begin{equation}
	\mathbf{f}_{knn}(\mathbf{x}') = \frac{1}{K} \sum_{i \in \mathcal{N}_K(\mathbf{x}')} \mathbf{y}_i
\end{equation}
with set $\mathcal{N}_K(\mathbf{x}')$ containing the indices of the $K$-nearest neighbors of $\mathbf{x}'$. The idea of KNN is based on the assumption of locality in data space: In local neighborhoods of $\mathbf{x}$ patterns are expected to have similar output values $\mathbf{y}$ (or class labels) to $\mathbf{f}(\mathbf{x})$. Consequently, for an unknown $\mathbf{x}'$ the label must be similar to the labels of the closest patterns, which is modeled by the average of the output value of the $K$ nearest samples. KNN has been proven well in various applications, e.g., in detection of quasars in interstellar data sets \cite{quasar}. %A survey of KNN variants has recently been given by Bhatia and Vandana \cite{knn}.

%The number $K$ of KNN has regularization capabilities: For $K=1$, KNN regression overfits to the output value of the nearest neighbor of $\mathbf{x}$, for $K=N$ it averages over all data samples.

% on the similarity of the concepts:

% 1. KNN regression is equivalent to the Nadaraya-Watson estimator \cite{nadaraya,watson} with a uniform kernel and a bandwidth that is chosen as the distance to the $K$-nearest point $h = \max_{i \in \mathcal{N}_K(\mathbf{x}')} \{\| \mathbf{y} - \mathbf{y}_i \| \}$....

% 2. the weighted distance approach is euquivalent to NW for triangle kernel

\section{Unsupervised KNN Regression}
\label{sec:UNN}

In this section we introduce two iterative strategies for UNN regression based on minimization of the data space reconstruction error (DSRE) \cite{Meinicke}.

\subsection{Unsupervised Regression}

Let $\mathbf{Y} = (\mathbf{y}_1, \ldots \mathbf{y}_N)$ with $\mathbf{y} \in \R^d$ be the matrix of high-dimensional patterns in data space. We seek for a low-dimensional representation, i.e., a matrix of latent points $\mathbf{X} = (\mathbf{x}_1, \ldots \mathbf{x}_N)$, such that a regression function $\mathbf{f}$ applied to $\mathbf{X}$ ,,point-wise optimally reconstructs the pattern'', i.e., we search for an $\mathbf{X}$ that minimizes
\begin{equation}
E(\textbf{X}) = \frac{1}{N} \| \textbf{Y} - \textbf{f}(\textbf{x};\textbf{X}) \|^2_F.
\label{dsre}
\end{equation}
$E(\textbf{X})$ is called data space reconstruction error (DSRE). Latent points $\mathbf{X}$ define the low-dimensional representation. The regression function applied to the latent points should optimally \textit{reconstruct} the high-dimensional patterns.

\subsection{UNN}

An UNN regression manifold is defined by variables $\mathbf{x} \in \textbf{X} \subset \R^q$ with the unsupervised formulation of an UNN regression manifold
\begin{equation}
	\label{eq:UNN}
	\textbf{f}_{UNN}(\textbf{x};\textbf{X}) = \frac{1}{K} \sum_{i \in \mathcal{N}_K(\mathbf{x},\textbf{X})} \mathbf{y}_i.
\end{equation}
Matrix $\textbf{X}$ contains the latent points $\mathbf{x}$ that define the manifold, i.e., the low-dimensional representation of data $\textbf{Y}$. Parameter $\textbf{x}$ is the location where the function is evaluated. An optimal UNN regression manifold minimizes the DSRE
\begin{equation}
E(\textbf{X}) = \frac{1}{N} \| \textbf{Y} - \textbf{f}_{UNN}(\textbf{x};\textbf{X}) \|^2_F,
\label{unndsre}
\end{equation}
with Frobenius norm
\begin{equation}
\|\mathbf{A}\|_F^2 = \sqrt{\sum_{i = 1}^d \sum_{j = 1}^N |a_{ij}|^2}.
\end{equation}
In other words: an optimal UNN manifold consists of low-dimensional points $\mathbf{X}$ that minimize the reconstruction of the data points $\mathbf{Y}$ w.r.t. KNN regression. Regularization in UNN regression may be not as important as regularization in other methods that fit into the unsupervised regression framework. For example, in UKR regularization means penalizing extension in latent space with $ E(\textbf{X})_p =  E(\textbf{X}) + \lambda \| \mathbf{X} \|$, and weight $\lambda$ \cite{ritter07}. In KNN regression moving the low-dimensional data samples infinitely apart from each other does not have the same effect as long as we can still determine the K-nearest neighbors, but extension can be penalized to avoid redundant solutions. For practical purposes (limitation of size of numbers) it might be reasonable to restrict continuous KNN latents spaces, e.g., to $\mathbf{x} \in [0,1]^q$. In the following section fixed latent space topologies are used that do not require further regularization.

\subsection{Iterative Strategy 1}
\label{sec:combinatorial}

For KNN not the absolute positions of data samples in latent space are relevant, but the relative positions that define the \textit{neighborhood relations}. This perspective reduces the problem to a combinatorial search for neighborhoods $\mathcal{N}_K(\mathbf{x}_i,\mathbf{X})$ with $i=1,\ldots,N$ that can be solved by testing all combinations of $K$-element subsets of $N$ elements, i.e., all $N \choose K$ combinations. The problem is still difficult to solve, in particular for high dimensions. In the following, we introduce a combinatorial approach to UNN, and introduce two iterative local strategies.

The idea of our first iterative strategy  (UNN~1) is to iteratively assign the data samples to a position in an existing latent space topology that leads to the lowest DSRE. We assume fixed neighborhood topologies with equidistant positions in latent space, and therefore restrict the optimization problem of Equation (\ref{eq:UNN}) to a search in a subset of latent space.

\begin{figure}[h]
\begin{center}
	\includegraphics[scale=.65]{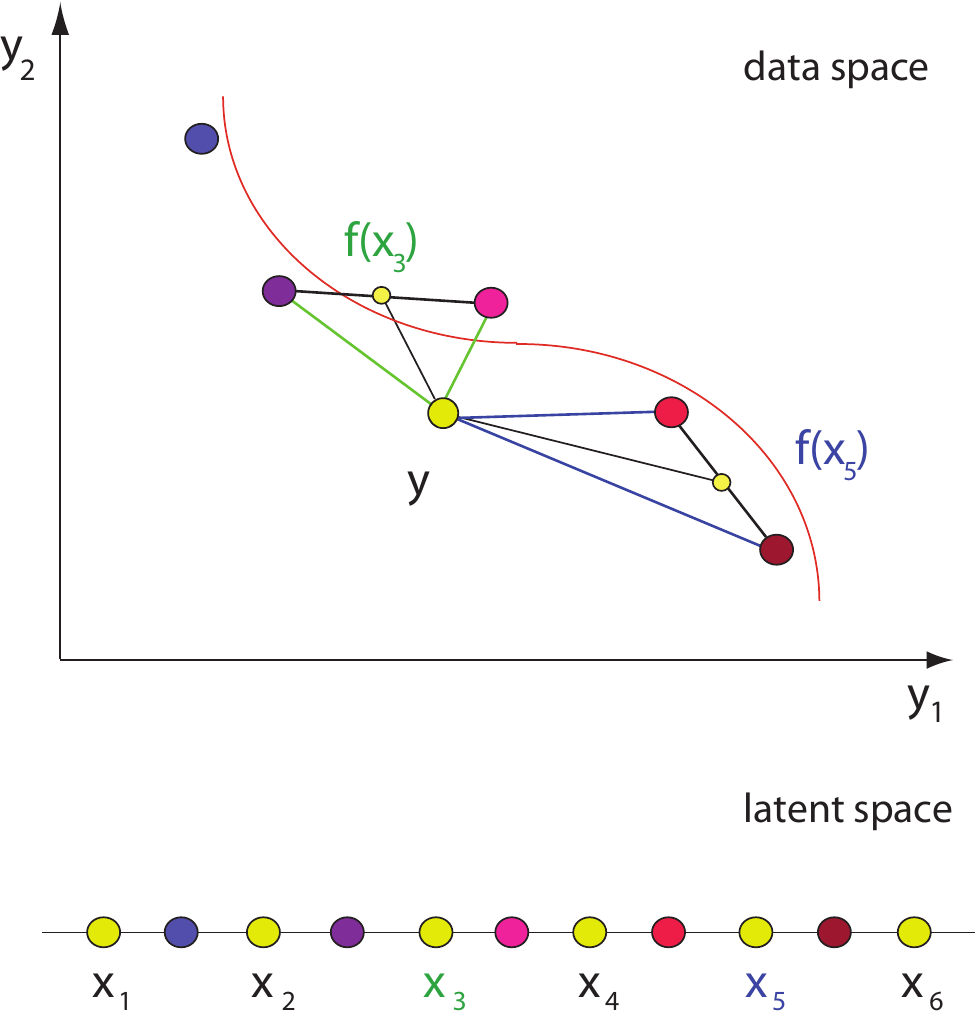}
\caption{\label{fig:local1}UNN~1: illustration of embedding of a low-dimensional point to a fixed latent space topology w.r.t. the DSRE testing all $\hat N+1$ positions.}
\end{center}
\end{figure}

As a simple variant we consider the linear case of the latent variables arranged equidistantly on a line $\mathbf{x} \in \R$. In this simplified case only the order of the elements is important. 
The first iterative strategy works as follows:
\begin{enumerate}
	\item Choose one element $\mathbf{y} \in \mathbf{Y}$,
	\item test all $\hat N+1$ intermediate positions of the $\hat N$ embedded elements in latent space,
%	\item choose the latent position that achieves the lowest DSRE, and embed $\mathbf{y}$,
	\item choose the latent position with $\min E(\textbf{X})$, and embed~$\mathbf{y}$,
	\item remove $\mathbf{y}$ from $\mathbf{Y}$, and repeat from Step 1 until all elements have been embedded.
\end{enumerate}
Figure \ref{fig:local1} illustrates the $\hat N+1$ possible embeddings of a data sample into an existing order of points in latent space (yellow/bright circles). For example, the position of element $\mathbf{x}_3$ results in a lower DSRE with $K=2$ than the position of $\mathbf{x}_5$, as the mean of the two nearest neighbors of $\mathbf{x}_3$ is closer to $\mathbf{y}$ than the mean of the two nearest neighbors of $\mathbf{x}_5$.

The complexity of UNN~1 can be described as follows. Each DSRE evaluation takes $K d$ computations. We assume that the $K$ nearest neighbors are saved in a list during the embedding for each latent point $\mathbf{x}$, so that the search for indices $\mathcal{N}_K(\mathbf{x},\textbf{X})$ takes $\mathcal{O}(1)$ time. The DSRE has to be computed for $N+1$ positions, which takes $(N+1) \cdot K d$ steps, i.e., $\mathcal{O}(N)$ time.

%- complexity: 
%	- - k * d is one DSRE computation
%	- UNN~1: N * (k * d)  z.B. 1000 * 10 * 100 = 1,000,000
%	- UNN~2: N * d + 2 * (k * d) z.B. 1000 * 100 + 2 * 10 * 100 = 100,000 + 2,000 = 102,000 

\subsection{Iterative Strategy 2}

\begin{figure}[h]
\begin{center}
		\includegraphics[scale=.65]{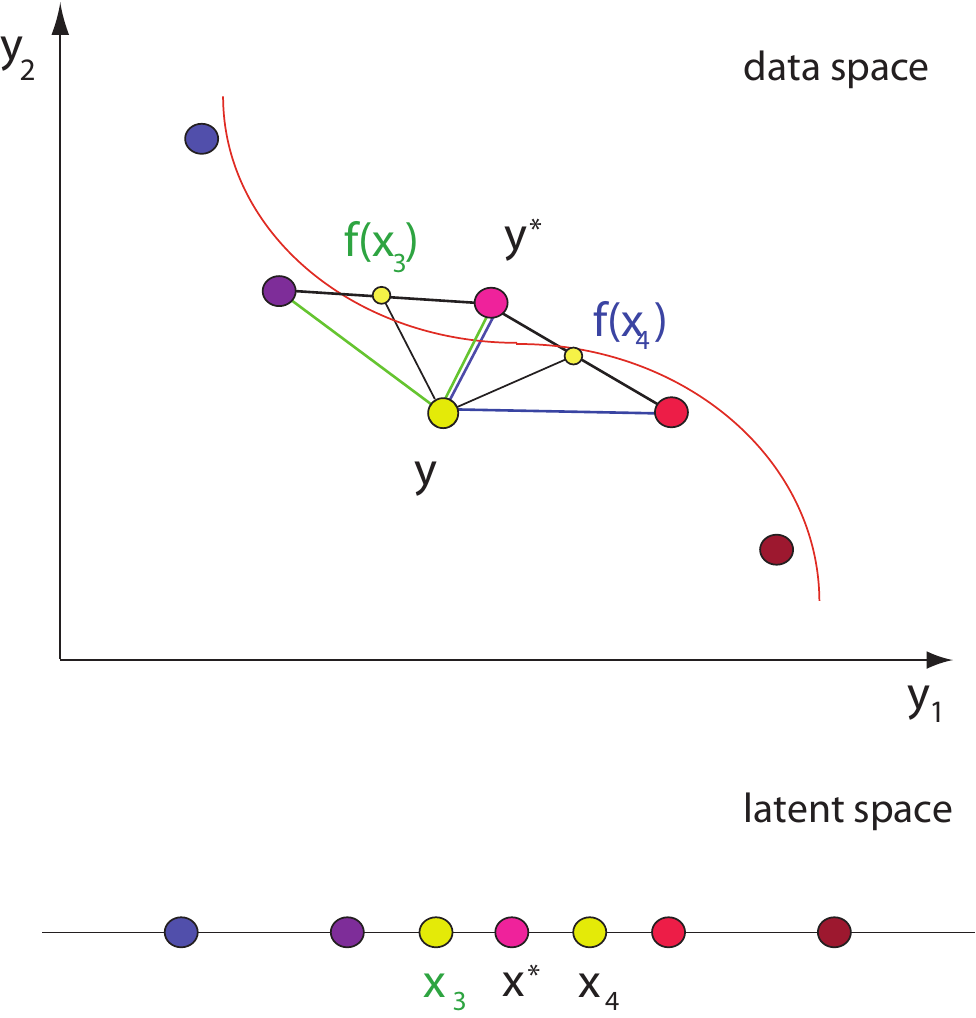}
\caption{\label{fig:local2}UNN~2: testing only the neighbored positions of the nearest point $\mathbf{y}^*$ in data space.}
\end{center}
\end{figure}

The iterative approach introduced in the last section tests all intermediate positions of previously embedded latent points. We propose a second iterative variant (UNN~2) that only tests the neighbored intermediate positions in latent space of the nearest embedded point $\mathbf{y}^* \in \hat{\mathbf{Y}}$ in data space. The second iterative strategy works as follows:
\begin{enumerate}
	\item Choose one element $\mathbf{y} \in \mathbf{Y}$,
	\item look for the nearest $\mathbf{y}^* \in \hat{\mathbf{Y}}$ that has already been embedded (w.r.t. distance measure like Euclidean distance),
%	\item choose the latent position next to $\mathbf{y}^*$ that achieves the lowest DSRE, and embed $\mathbf{y}$,
	\item choose the latent position next to $\mathbf{y}^*$ with $\min E(\textbf{X})$ and embed $\mathbf{y}$,
	\item remove $\mathbf{y}$ from $\mathbf{Y}$, add $\mathbf{y}$ to $\hat{\mathbf{Y}}$, and repeat from Step 1 until all elements have been embedded.
\end{enumerate}
Figure \ref{fig:local2} illustrates the embedding of a 2-dimensional point $\mathbf{y}$ (yellow) left or right of the nearest point $\mathbf{y}^*$ in data space. The position with the lowest DSRE is chosen. In comparison to UNN~1, $\hat N$ distance comparisons in data space have to be computed, but only 2 positions have to be tested w.r.t. the data space reconstruction error.
UNN~2 computes the nearest embedded point~$\mathbf{y}^*$ for each data point, which takes $Nd$ steps. Only for the two neighbors the DSRE has to be computed, resulting in an overall number of $Nd+ 2 Kd$ steps, i.e., it takes $\mathcal{O}(N)$ time. Because of the multiplicative constants, UNN~2 is faster in practice. For example, for $N=1,000$, $K=10$, and $d=100$, UNN~1 takes $1,001,000$ steps, while UNN~2 takes $102,000$ steps. Testing all combinations takes $1000 \choose 10$ steps, which is not computable in reasonable time. The following experimental section will answer the question, if this speedup of UNN~2 has to be paid with worse DSREs.

\subsection{Experiments}

This section shows the behavior of the iterative strategies on three test problems. We will compare the DSRE of both strategies to the initial DSRE at the end of this section.

\subsubsection{2D-$S$}

\begin{figure}[h]
\begin{center}
\subfigure[]{\includegraphics[scale=.215]{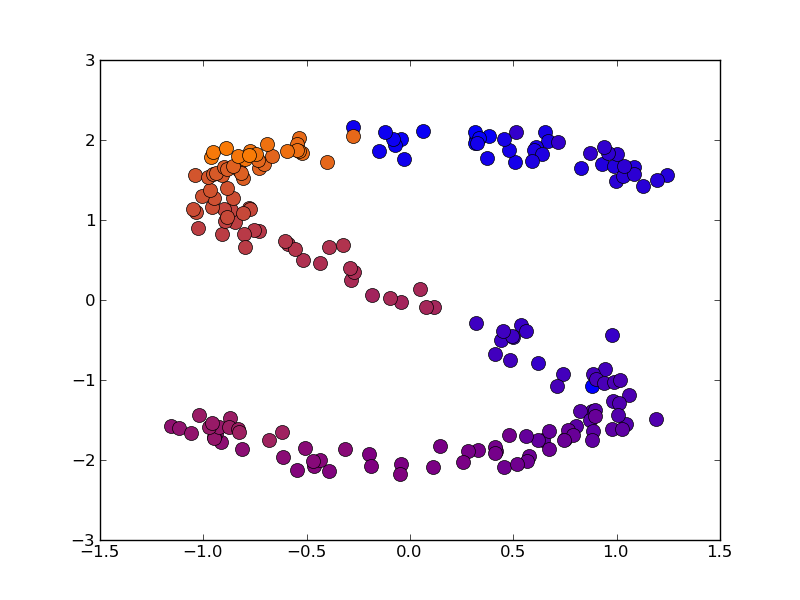}}
\subfigure[]{\includegraphics[scale=.215]{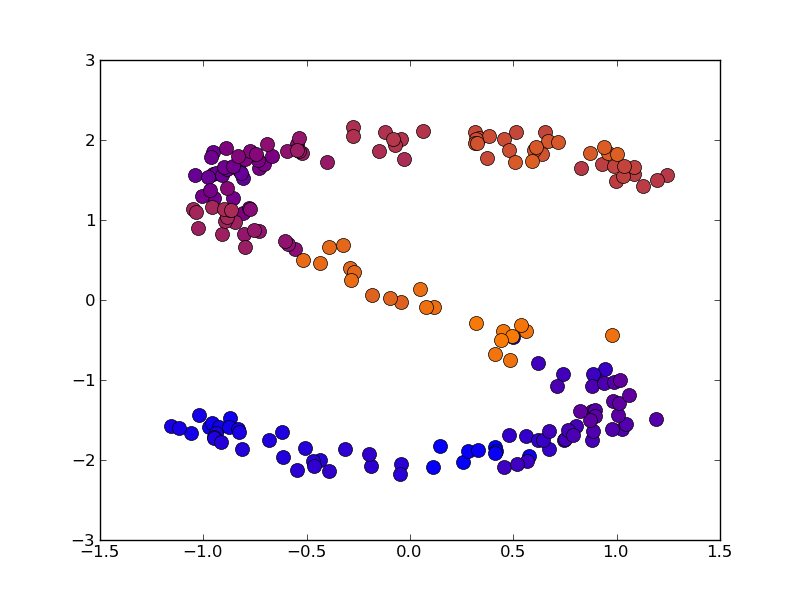}}
%\subfigure[]{\includegraphics[scale=.21]{pics/spiral1.png}}
%\subfigure[]{\includegraphics[scale=.21]{pics/spiral2.png}}					
\caption{\label{fig:2d}(a) UNN~1, and (b) UNN~2 embedding with $K=5$ on 2D-$S$.}
\end{center}
\end{figure}

First, we compare UNN~1 and UNN~2 on a simple 2-dimensional data set, i.e., the 2-dimensional noisy $S$ with $N=200$ (2D-$S$). Figure \ref{fig:2d} shows the experimental results with $K=5$ nearest neighbors. Similar colors correspond to neighbored latent
points. Part~(a) shows an UNN~1 embedding of the 2D-$S$ data set. Part~(b) shows the embedding of the same data set with UNN~2. The colors of both embeddings show a satisfying topological sorting, although we can observe local optima.

\subsubsection{3D-$S$}

\begin{figure}[b]
\begin{center}
\subfigure[]{\label{fig:3d-a}\includegraphics[scale=.22]{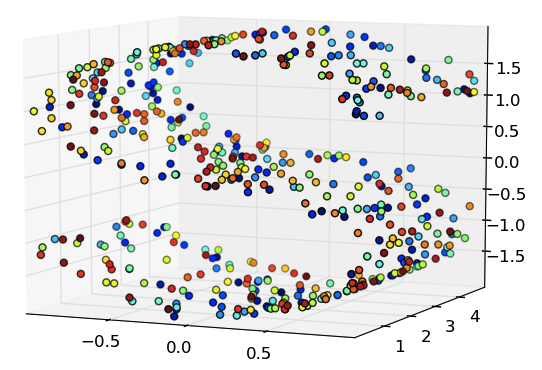}}
\subfigure[]{\includegraphics[scale=.22]{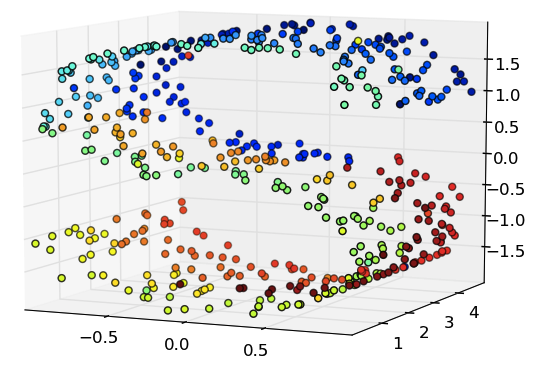}}
\subfigure[]{\includegraphics[scale=.22]{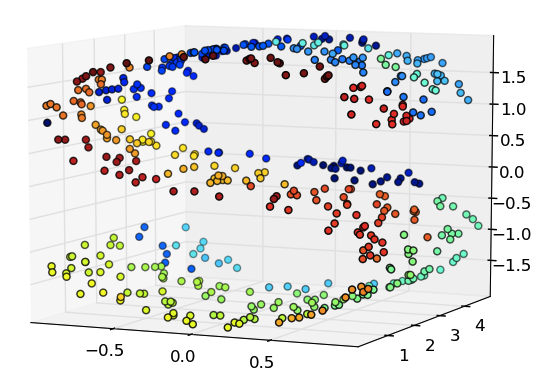}}
\subfigure[]{\includegraphics[scale=.22]{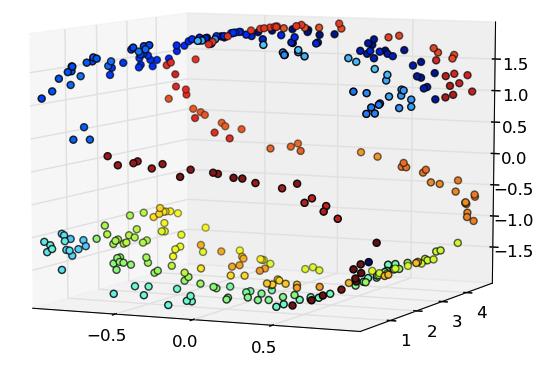}}	
\caption{\label{fig:3d} Results of UNN on 3D-$S$: (a) the unsorted $S$ at the beginning, (b) the embedded $S$ with UNN~1 and $K=10$, (c) the embedded $S$ with UNN~2 and $K=10$, and (d) a variant of $S$ with a hole embedded with UNN~2.}
\end{center}
\end{figure}

In the following, we will test UNN regression on a 3-dimensional $S$ data set (3D-$S$). The variant without a hole consists of 500 data points, the variant with a hole in the middle consists of 400 points. Figure \ref{fig:3d} (a) shows the order of elements of the 3D-$S$ data set 
without a hole at the beginning. The corresponding embedding with UNN~1 and $K=10$ is shown in Part~(b) of the figure. Again, similar colors correspond to neighbored points in latent space. Part~(c) of Figure \ref{fig:3d} shows the UNN~2 embedding achieving similar results. Also on the UNN embedding of the $S$ data set with hole, see Part~(d) of the figure, a reasonable neighbored assignments can be observed. Quantitative results for the DSRE are reported in Table \ref{tab:dsre}.

\subsubsection{USPS Digits}

Last, we experimentally test UNN regression on test problems from the USPS digits data set \cite{usps}. For this sake we take 100 data samples of 256-dimensional (16 x 16 pixels) pictures of handwritten digits of 2's and 5's. We embed a one-dimensional manifold, and show the 
\begin{figure}[h!]
\begin{center}

\subfigure[]{	\hspace{-0.4cm}\includegraphics[scale=.18]{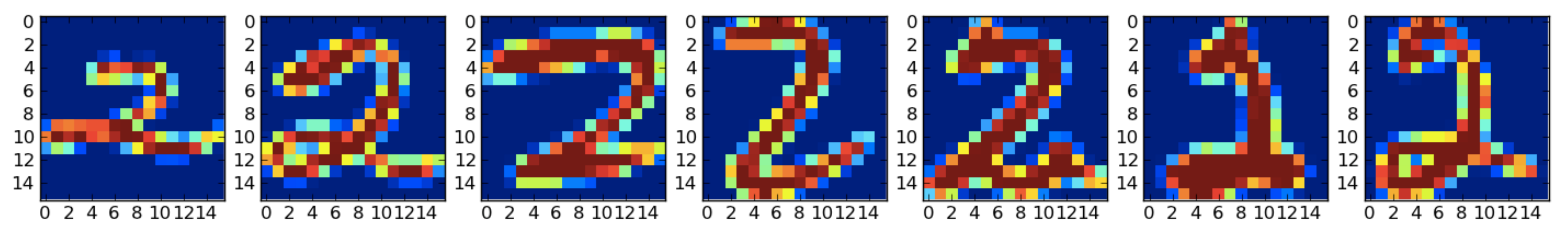}}
%	\hspace{-0.4cm}
\subfigure[]{	\hspace{-0.4cm}\includegraphics[scale=.18]{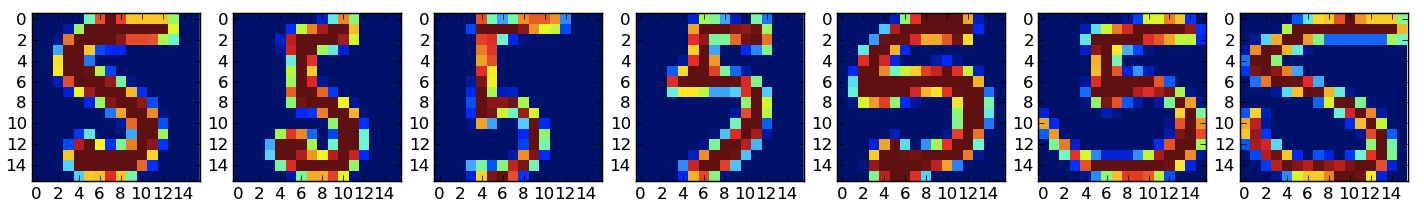}}					
\caption{\label{fig:7up}UNN~2 embeddings of USPPS digits: (a) 2's, and (b) 5's. Digits are shown that are assigned to every 14th embedded latent point. Similar digits are neighbored in latent space.}
\end{center}
\end{figure}
high-dimensional data that is assigned to every 14th latent point, i.e., neighbored digits in the plot are neighbored in latent space. Figure \ref{fig:7up} shows the result of the UNN~2-embedding for 2's and 5's with $K=10$. We can observe that neighbored digits are similar to each other, while digits that are dissimilar are further away from each other in latent space.

\subsubsection{DSRE Comparison}
\label{DSRE Comparison}

Last, we compare the DSRE achieved by both strategies with the initial DSRE, and the DSRE achieved by LLE on all test problems. 
\begin{table}[b]
\newcommand{\rotated}[1]{\parbox[c]{3mm}{\rotatebox{45}{#1}}}
\caption{Comparison of DSRE for initial data set, and after embedding with strategy UNN~1, and UNN~2\label{tab:dsre}.}
\small
\begin{center}%
	\begin{tabular}{l|rrr|rrrrr}
			& \multicolumn{3}{c|}{2D-$S$} &  \multicolumn{3}{c}{3D-$S$}\\
	\parbox{1.2cm}{$K$} &  2 & 5 & 10 &  2 & 5 & 10 \\
\hline
init&  	201.6 & 	290.0  &  	309.2  & 691.3  &  904.5& 945.80 \\
UNN~1 &  	\textbf{19.6}& 		\textbf{27.1} & 		66.3 &  \textbf{101.9} & \textbf{126.7} & \textbf{263.39}\\
UNN~2 & 	29.2  & 	70.1  &  	64.7 & 140.4 & 244.4 & 296.5 \\
LLE & 25.5 & 37.7 & \textbf{40.6} & 135.0& 514.3 & 583.6\\
\hline
		& \multicolumn{3}{c|}{3D-$S$ hole} &  \multicolumn{3}{c}{digits (7)}\\
	\parbox{1.2cm}{$K$} &  2 & 5 & 10 &  2 & 5 & 10  \\
\hline
init&  	  577.0 & 727.6 & 810.7 & 196.6 & 248.2 & 265.2  \\
UNN~1 &  \textbf{80.7}	& \textbf{108.1} & \textbf{216.4} & \textbf{139.0} &  \textbf{179.3} &  \textbf{216.6} \\
UNN~2 & 101.8  & 204.4 & 346.8  & 145.3 & 195.4 & 222.1 \\
LLE & 94.9 & 198.9 & 387.4 & 147.8 & 198.1& 217.8  \\
\end{tabular}
\end{center}
\end{table}
For the USPS digits data set we choose the number $7$. Table \ref{tab:dsre} shows the experimental results of three settings for the neighborhood size $K$. The lowest DSRE on each problem is highlighted with bold figures. After application of the iterative strategies the DSRE is significantly lower than initially. Increasing $K$ results in higher DSREs. With exception of LLE with $K=10$ on 2D-$S$, the UNN~1 strategy always achieves the best results. UNN~1 achieves lower DSREs than UNN~2, with exception of 2D-$S$, and $K=10$. The win in accuracy has to be paid with a constant runtime factor that may play an important role in case of large data sets, or high data space dimensions.

% 3d no hole
% 2: DSRE 1: 694.913446221DSRE 2: 101.991374121
% 5: DSRE 1: 874.606395525DSRE 2: 126.730671097
% 10: DSRE 1: 946.405962791DSRE 2: 263.390061058

% 3d hole
% 2: DSRE 1: 617.245873084DSRE 2: 80.7760145875
% 5: DSRE 1: 737.809256129DSRE 2: 108.132535812
% 10 DSRE 1: 828.721733673DSRE 2: 216.494118705

% 7
% 2 DSRE 1: 195.894428676DSRE 2: 139.059858886
% 5 DSRE 1: 246.933183912DSRE 2: 179.357399687
% 10 DSRE 1: 262.217230293DSRE 2: 216.662636113

%to do
%- kombinatorisch: nur K-elementigen teilmengen zuordnen, n über k
%- mehrmals sortieren! bringt evtl verbesserung, testen!

%1. andere ziffer!
%2. UNN~1 DSRE tabelle
%3. weiteres 3d-S

\section{Conclusions}
\label{sec:conclusions}

%Fast dimensionality reduction methods are required that are able to process huge data sets, and large dimensions.
With UNN regression we have fitted a fast regression technique into the unsupervised setting for dimensionality reduction. The two iterative UNN strategies are efficient methods to embed high-dimensional data into fixed one-dimensional latent space taking $\mathcal{O}(N)$ time. The speedup is achieved by restricting the number of possible solutions (reduction of solution space), and applying fast iterative heuristics. Both methods turned out to be performant on test problems in first experimental analyses. UNN~1 achieves lower DSREs, but UNN~2 is slightly faster because of the multiplicative constants of UNN~1. Our future work will concentrate on the analysis of local optima the UNN strategies approximate, and how the approach can be extended to guarantee global optimal solutions. Furthermore, the UNN strategies can be extended to latent topologies with higher dimensionality. For $q=2$ the insertion of intermediate solutions into a grid is more difficult: it results in shifting rows and columns of the grid, and thus changes the latent topology in parts that may not be desired. A simple stochastic search strategy can be employed that randomly swaps positions of latent points in the grid.

% Another possible extension of UNN is a continuous backward mapping from latent to data space $\mathbf{f}: \mathbf{x} \rightarrow \mathbf{y}$ employing a distance-weighted variant of KNN.

\vspace{0.5cm}

\bibliographystyle{abbrv}

\end{document}